\documentclass[10pt,journal,compsoc,two column]{IEEEtran}%

\usepackage[utf8]{inputenc} 
\usepackage[T1]{fontenc}    
\usepackage{hyperref}       
\usepackage{url}            
\usepackage{booktabs}       
\usepackage{amsfonts}       
\usepackage{nicefrac}       
\usepackage{microtype}      
\usepackage{xcolor}         
\usepackage{amsmath}
\usepackage{amssymb}
\usepackage{graphicx}
\usepackage{multirow}
\usepackage{epstopdf}
\usepackage[ruled,longend]{algorithm2e}
\usepackage{subcaption}
\usepackage{pdfpages}

\title{Age-Invariant Face Embedding using the Wasserstein Distance}

\begin{document}

\author{~Eran~Dahan and Yosi~Keller~
\IEEEcompsocitemizethanks{\IEEEcompsocthanksitem  E. Dahan \& Y. Keller are with the Faculty of Engineering, Bar-Ilan University,
Email:yosi.keller@gmail.com}\thanks{Manuscript received April 19, 2005;
revised August 26, 2015.}}
\maketitle

\begin{abstract}
 In this work, we study face verification in datasets where images of the same individuals exhibit significant age differences. This poses a major challenge for current face recognition and verification techniques. To address this issue, we propose a novel approach that utilizes multitask learning and a Wasserstein distance discriminator to disentangle age and identity embeddings of facial images. Our approach employs
multitask learning with a Wasserstein distance discriminator that minimizes the mutual information between the age and identity embeddings
by minimizing   the Jensen-Shannon divergence. This improves the encoding of age and identity information in face images and enhances the performance of face verification in age-variant datasets. We evaluate the effectiveness of our approach using multiple age-variant face datasets and demonstrate its superiority over state-of-the-art methods in terms of face verification accuracy.
\end{abstract}

\section{Introduction}

Face recognition and verification are fundamental tasks in computer vision,
with a wide range of applications in security, surveillance, and
human-computer interaction. A number of techniques have been
suggested enhancing the accuracy of facial recognition, including the
computation of face features through learning \cite{rep1, rep2}, as
well as the use of deep learning methods \cite{DEEPID, deepface} and
training losses such as the Triplet \cite{vgg_triplet_loss_ref} and Large
Margin losses \cite{deng2018arcface,Cosface,elastic}. Although these methods have been
effective, changes in appearance caused by age, pose, and
lighting conditions can significantly decrease their accuracy. Therefore,
facial recognition systems strive to increase the separation between different
identities (interclass separation) while decreasing the separation
between images of the same identity (intraclass separation).
Age-invariant face verification is a crucial component of some facial
recognition tasks, with many applications, including locating missing
children \cite{lostchild, 9411913}, medical uses \cite{medicine, medicine2},
kinship verification \cite{kinship, kinship2}, and demographic estimation
\cite{demographic}, among others. However, the age-related variations in
facial appearance (see Fig. 1 of the Appendix) might be significant as the age
gap between the images increases. For instance, using face images taken
decades earlier for face verification, decreases the verification accuracy in
real-world datasets. Various methods have been proposed to address
age-invariant face verification, which can be broadly categorized as
intraclass minimization, face synthesis, and disentanglement methods.
\begin{figure}[t]
\begin{center}
\centering\includegraphics[width=1.0\linewidth]{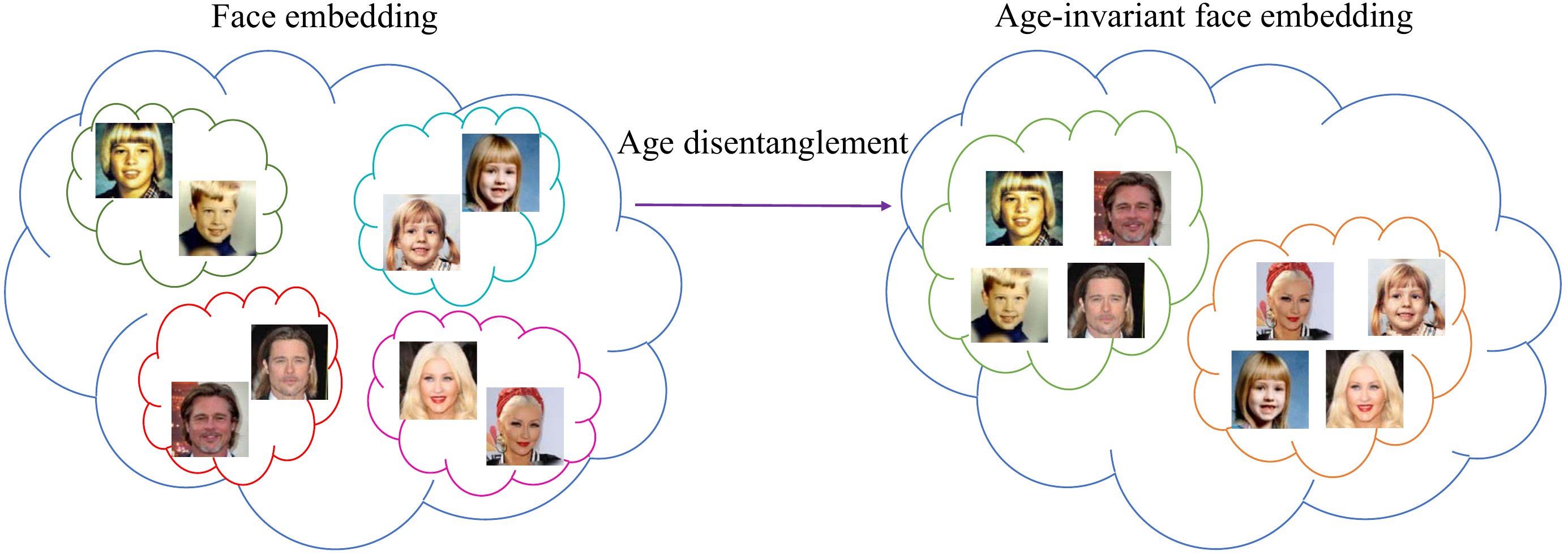}
\end{center}
\caption{Age and identity disentanglement. The left-hand side shows the common face embedding space that encodes both age and identity, with the embeddings of each person grouped separately by age. The right-hand side shows an age-invariant embedding, in which all images of the same person are grouped together, regardless of their substantial age differences.}%
\label{fig:main_idea}%
\end{figure}
Face synthesis methods synthesize face images of the same subjects
(identities) at different ages to enlarge the training set per identity and
reflect the aging effect \cite{9411913, 9146699, learning}. These methods are
based on using generative adversarial networks (GANs). However, although GANs
have made significant improvements in recent years, such methods only enlarge
the training dataset and do not improve the image embeddings.
Recent methods for age-invariant face verification apply statistical
disentanglement to disentangle the biometric attributes, as shown
in Fig. \ref{fig:main_idea}. In most cases, as in age disentanglement, the
attribute we wish to disentangle from the image embedding is not linearly
separable from the other biometric face attributes. Given a face image
$\boldsymbol{x}$, its embedding $\widehat{\boldsymbol{x}},$ with identity and
age attributes $\boldsymbol{x_{id}}$, $\boldsymbol{x_{age}}$, respectively,
there is no straightforward analytical formulation to separate
$\boldsymbol{x_{id}}$ or $\boldsymbol{x_{age}}$ from $\widehat{\boldsymbol{x}%
}$. Additionally, it is difficult to model the relationship between these two
attributes, requiring statistical disentanglement methods. Thus, various
disentanglement methods have been proposed, including probabilistic models
\cite{Gong}, Expectation Maximization (EM) techniques \cite{Gong_new}, and
gradient orientation pyramids \cite{Ling}.

In this work, we propose to improve the statistical disentanglement of the age
and identity attributes of face images. We derive an age-invariant
identity embedding $\boldsymbol{x_{id}}$ that is
invariant to the age attribute $\boldsymbol{x_{age}}$. For that, we propose a
novel method to disentangle the age and identity features with respect to
their mutual information. Statistical disentanglement was first studied by Independent Component Analysis (ICA) \cite{GEPSHTEIN2019368,fastica} optimizing statistical divergence measures between the estimated signals. Recent work by Huo et al. \cite{Hou} and Xie et al.
\cite{9682736} also considered minimizing mutual information to disentangle
age and identity through neural estimation. In contrast, our approach utilizes
a multitask learning architecture with a Wasserstein distance discriminator that
minimizes the Jensen-Shannon divergence. We denote the resulting age-invariant
face image embedding as the Wasserstein Mutual Information for Age Invariant
(WMI-AI). Our method is shown to effectively reduce the mutual information
between age and identity embeddings, and is similar to the use
of the Wasserstein distance as a discriminator to stabilize WGAN training
\cite{WGAN}, or as a discriminator for domain adaptation \cite{WDGR}.

In particular, we propose the following contributions:

\begin{itemize}
\item We propose a multitask learning architecture that disentangles the age
and identity embeddings while optimizing the embeddings for age classification
and face recognition.

\item We suggest a discriminator based on the Wasserstein distance and
Jensen-Shannon divergence to minimize the mutual information between the
embedding tasks.

\item Through extensive evaluation of major cross-age face verification
datasets, including CALFW \cite{CALFW}, AgeDB30 \cite{agedb30}, CACD-VS \cite{cacdvs}, and ECAF \cite{learning}, we demonstrate that our method achieves state-of-the-art accuracy, even when the age gap between the pair of compared images is significant.
\end{itemize}


\section{Related Work}

\label{Related_Work}

\subsection{Age-Invariant Face Image Embedding}
\textbf{Learning the aging patterns in face images.} Some researchers have
focused on learning the manifestation of the aging process in face images.
Park et al. \cite{Park} proposed a 3D face age model to compensate for age
variations in 2D probe face images. Ramanathan et al. \cite{Ramanathan}
suggested a shape transformation model to model the aging process. However,
these methods require large annotated datasets to achieve high performance.

\textbf{Face synthesis}. Generative adversarial networks (GANs) were used to
synthesize face images with varying ages. Debayan et al. \cite{9411913}
proposed synthesizing a face image with a known identity at a specific age and
then comparing it with a query image using a face verification scheme. Zhao et
al. \cite{9146699} proposed an end-to-end training architecture for face
synthesis and recognition, while Huang et al. \cite{learning} proposed a
multitask learning framework for face recognition, age classification, and
face synthesis.

\textbf{Age disentanglement}. Another line of research focuses on modeling the
face image embedding as a nonlinear combination of identity and age-related
factors, and disentangling the two. Gong et al. \cite{Gong} proposed a
probabilistic model comprising a latent identity factor that is age-invariant
and a latent age factor. The model was trained using expectation-maximization
(EM). Gong et al. \cite{Gong_new} later proposed an entropy maximization
descriptor that proved to be more discriminative and resulted in improved
verification accuracy using the age-invariant features. Wen et al. \cite{7780898} suggested
a deep learning framework to learn the latent age factor, while SVM with a
gradient orientation pyramid (GOP) was proposed by Ling et al. \cite{Ling} to
discard the age information from an image. In contrast to these works, which
factorize the face image, our WMI-AI approach disentangles the age from
identity features by learning a mutual representation that minimizes their
mutual information while learning age-invariant face embeddings for recognition.

\textbf{Discriminative convolutional neural networks}. Recent research on
age-invariant face verification has focused on improving the recognition
models through discriminative approaches. Wang et al. \cite{Wang1} proposed
the OE-CNN (Orthogonal-Embedding Convolutional Neural Network), which learns
an orthogonal representation for the embedding of the face and age using
angular and radial representations. Another approach, proposed by Wang et al.
\cite{8954207}, uses a decorrelation method to decompose age and identity
features through decorrelated adversarial learning (DAL). Lee et al.
\cite{Lee2021ImprovingFR} proposed to improve verification results for child
images using an inter-prototype loss function that minimizes the similarity of
child images and resulted in improved verification accuracy of adult-child
image pairs.

\subsection{Mutual Information in Representation Learning}

The use of mutual information (MI) in representation learning has gained
increasing attention in recent years. Schwartz and Tishby
\cite{shwartz2017opening} used MI to analyze the relationship
between the different layers of a deep neural network (DNN). Later, Tishby and
Zaslavsky \cite{7133169} applied the bottleneck principle to optimize the size
of the representation in a DNN through mutual information minimization. As
MI is difficult to compute, researchers have focused on using
convolutional neural networks (CNNs) to estimate it. Cheng et al.
\cite{cheng2020club} proposed CLUB, which uses a CNN to estimate the MI between samples from an unknown distribution using a lower
contrastive logarithmic ratio bound. Belghazi et al. \cite{belghazi2018mutual}
introduced MINE - mutual information neural estimation, which computes a lower
bound of the MI using a critic DNN network, and showed it to be a tight bound in various cases, including domain adaptation.

Mutual information also plays a role in age-invariant face recognition models.
Hou et al. \cite{Hou} and Xie et al. \cite{9682736} proposed to minimize
MI between the age and identity components of a face image.
These recent works mainly focus on using neural estimation for MI. In contrast, the proposed WMI-AI method minimizes the Wasserstein
distance of the Jensen-Shannon divergence to minimize the MI between the age and identity embeddings.

\subsection{Wasserstein Distance}

The Wasserstein distance is a measure of the distance between two probability
distributions and is widely used in the training of deep learning networks.
This distance metric has been particularly effective in the field of
generative adversarial networks (GANs), where it has helped stabilize
adversarial training \cite{WGAN}. Shen et al. \cite{WDGR} suggested the WD-GRL
Wasserstein Distance Guided Representation Learning for domain adaptation
applications, by estimating and minimizing the cross-domain Wasserstein
distance. The Wasserstein distance has also been applied to a range of
computer vision and machine learning tasks, including object detection
\cite{object}, scene classification \cite{scene}, and multi-source domain
adaptation \cite{multi-source} to name a few. In contrast, our approach involves the reduction of mutual information between age and identity embeddings by minimizing the Wasserstein distance. This aligns with our objective of disentangling these attributes to enhance the accuracy of face verification in the proposed WMI-AI scheme, particularly when analyzing face images that exhibit notable age discrepancies.
\section{Age-Invariant Face Embedding}
\label{Proposed Method}
\begin{figure*}[tbh]
\begin{center}
\centering\includegraphics[width=1.0\linewidth]{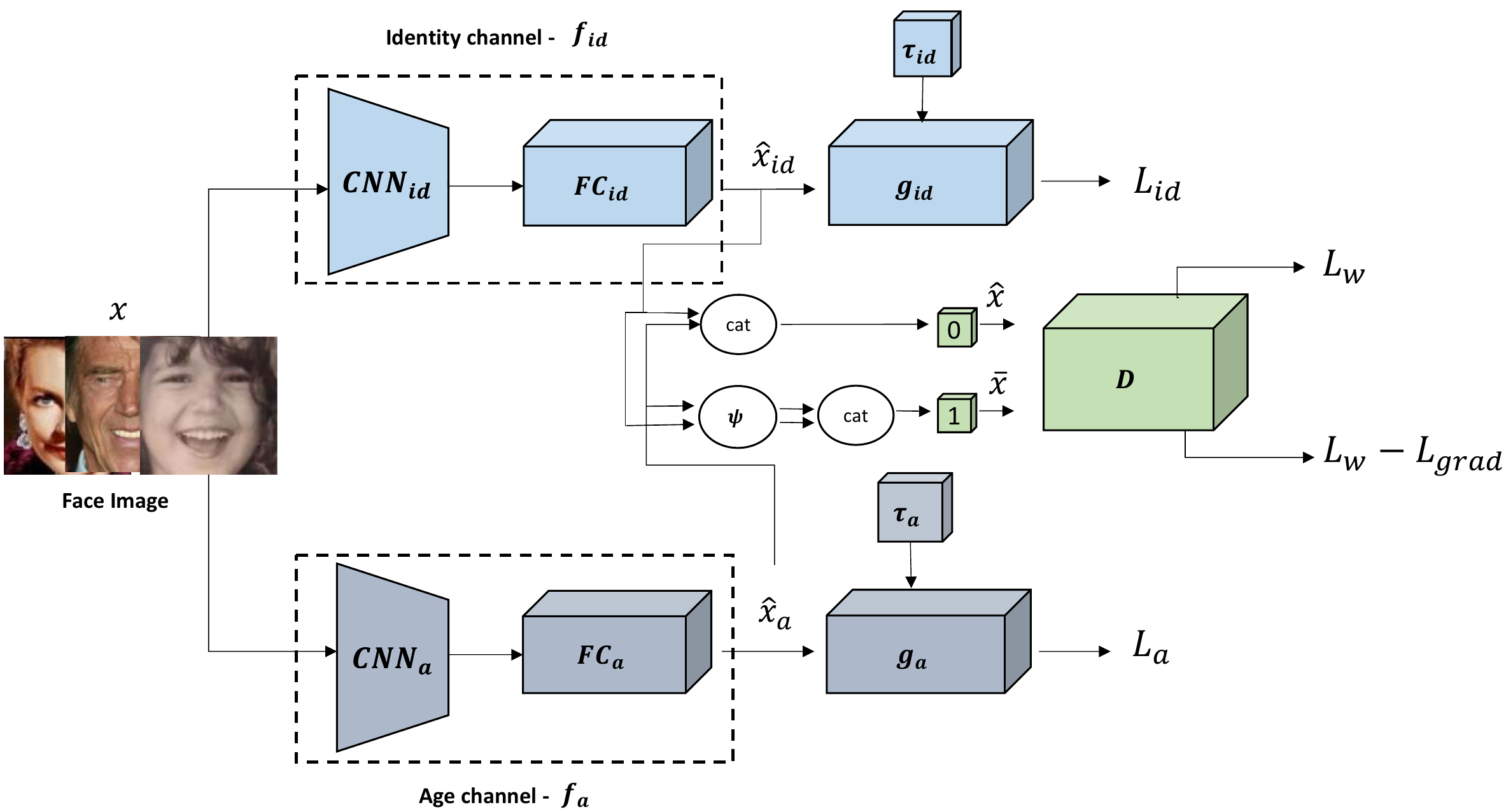}
\end{center}
\caption{The proposed network involves an identity encoder network
$\boldsymbol{f_{id}}$ and an age encoder $\boldsymbol{f_{a}}$. When given a
facial image $\boldsymbol{x}$, $\boldsymbol{f_{id}}$ generates the identity
embedding $\widehat{\boldsymbol{x}}\boldsymbol{_{id}}$, while
$\boldsymbol{f_{a}}$ computes the age embedding $\widehat{\boldsymbol{x}%
}\boldsymbol{_{a}}$. These encoders are optimized for facial recognition and
age estimation using the classifiers $\boldsymbol{g_{id}}$ and
$\boldsymbol{g_{a}}$. During adversarial training, the discriminator
$\boldsymbol{D}$ utilizes the embedding pairs $\boldsymbol{\hat{x}%
=}\left(  \widehat{\boldsymbol{x}}\boldsymbol{_{id}^{i},\widehat
{\boldsymbol{x}}_{a}^{i}}\right)  $ and randomly shuffled embedding pairs
$\boldsymbol{\bar{x}=}\left(  \widehat{\boldsymbol{x}}\boldsymbol{_{id}%
^{i},\widehat{\boldsymbol{x}}_{a}^{j}}\right)  $, $i\neq j,$ as positive and
negative samples, correspondingly.}%
\label{fig:full-arc}%
\vspace{-10pt}
\end{figure*}

In this work, we propose to train an age-invariant face embedding for face
verification. We propose to disentangle the age and identity attributes in a
high-dimensional embedding space by minimizing the Jensen-Shannon Divergence
(JSD) between the identity and age embeddings, while jointly optimizing the embeddings for face recognition and age estimation. In line
with Fig. \ref{fig:full-arc}, the proposed method consists of two components:
computing the face image embeddings for age and identity, and a mutual
information discriminator. Given a face image $\boldsymbol{x}$, we train the
age and identity embeddings, consisting of an identity CNN backbone and encoder $\boldsymbol{f_{id}}$ and an age CNN and encoder $\boldsymbol{f_{a}}$,
respectively.
The identity encoder outputs the embedding $\boldsymbol{\hat{x}_{id}}$, which
is passed through the identity classifier $\boldsymbol{g_{id}.}$ Similarly,
the age encoder outputs the embedding $\boldsymbol{\hat{x}_{a}}$, which is
then passed through a regression layer $\boldsymbol{g_{a}}$ to predict the
face image age. The face and age encoders are detailed in Section
\ref{sec:enc}. The MI discriminator $\boldsymbol{D}$ is used
to classify whether the mutual representation $\boldsymbol{\hat{x}_{id}%
},\boldsymbol{\hat{x}_{a}}$ is drawn from the mutual probability
$\mathbb{P}(\boldsymbol{\hat{x}_{id}},\boldsymbol{\hat{x}_{a}})$ or from the
independent probabilities $\mathbb{P}(\boldsymbol{\hat{x}_{id}}%
,\boldsymbol{\hat{x}_{a}})=\mathbb{P}(\boldsymbol{\hat{x}_{id}})\mathbb{P}%
(\boldsymbol{\hat{x}_{a}})$. The discriminator and its objective are detailed
in Section \ref{subsec:Discriminator}.

We train the embedding $\boldsymbol{\hat{x}_{id}}$ adversarially with the
discriminator loss, to fool the discriminator to predict that
$\boldsymbol{\hat{x}_{id}}$ and $\boldsymbol{\hat{x}_{a}}$ were drawn from an
independent distribution. However, adversarial training is known to be
unstable and difficult to converge, so we propose training the discriminator
using the Wasserstein distance cost, which was shown to improve the convergence
of adversarial training \cite{WGAN}. The multitask training and optimization
are described in Section \ref{sec:MLO}. Lastly, we discuss in Section
\ref{sec:UGT} the use of a pretrained age estimation network when the age
labels of the training set are unknown.

\subsection{Discriminator Architecture and Training Method}

\label{subsec:Discriminator}

We represent the mutual embedding of age $\boldsymbol{\hat{x}_{a}}%
\in\mathbb{R}^{d_{a}}$ and identity $\boldsymbol{\hat{x}_{id}}\in
\mathbb{R}^{d_{id}}$ as the random variable $\boldsymbol{\hat{x}}%
\triangleq(\boldsymbol{\hat{x}_{id}},\boldsymbol{\hat{x}_{a}})\in
\mathbb{R}^{d_{id}+d_{a}}$. Let $\boldsymbol{z}$ be a binary
random variable%
\begin{equation}
\boldsymbol{\hat{x}}\sim%
\begin{cases}
\mathbb{P}(\boldsymbol{\hat{x}_{id}},\boldsymbol{\hat{x}_{a}}), & \text{for
}\boldsymbol{z}=0\\
\mathbb{P}(\boldsymbol{\hat{x}_{id}})\mathbb{P}(\boldsymbol{\hat{x}_{a}}), &
\text{for }\boldsymbol{z}=1
\end{cases}
.
\label{equ:z}
\end{equation}
The mutual information between $\boldsymbol{\hat{x}}$ and $\boldsymbol{z}$ can
be expressed as:
\begin{equation}
I(\boldsymbol{\hat{x}};\boldsymbol{z})=JSD(\mathbb{P}(\boldsymbol{\hat{x}%
_{id}},\boldsymbol{\hat{x}_{a}})\,\,\,||\,\,\,\mathbb{P}(\boldsymbol{\hat
{x}_{id}})\mathbb{P}(\boldsymbol{\hat{x}_{a}}))
\end{equation}
where $JSD$ is the Jensen-Shannon Divergence. The objective of the
discriminator $\boldsymbol{D}$ is to distinguish between the two
representations in Eq. \ref{equ:z} and estimate the random variable
$\boldsymbol{z}$. Hence, the objective of $\boldsymbol{D}$ is to maximize the
$JSD$, while our adversarial-training objective is to minimize the $JSD$ with
respect to $\boldsymbol{\hat{x}_{id}}$, given $\boldsymbol{\hat{x}_{a}}$. The discriminator loss function is
thus given by:%
\begin{equation}%
L_{d}(\boldsymbol{\hat{x}})=  \frac{1}{2}{E}_{\boldsymbol{\hat{x}}%
\sim\mathbb{P}(\boldsymbol{\hat{x}_{id}},\boldsymbol{\hat{x}_{a})}}%
\log(1-D(\boldsymbol{\hat{x}}))+  \frac{1}{2}{E}_{\boldsymbol{\hat{x}}\sim\mathbb{P}(\boldsymbol{\hat{x}%
})\mathbb{P}(\boldsymbol{\hat{x}_{a}})}\log D(\boldsymbol{\hat{x}})
. \label{JSD_l}%
\end{equation}
Minimizing the $JSD$ between the distributions $\mathbb{P}(\boldsymbol{\hat
{x}_{id}},\boldsymbol{\hat{x}_{a}})$ and $\mathbb{P}(\boldsymbol{\hat{x}%
})\mathbb{P}(\boldsymbol{\hat{x}_{a}})$ is equivalent to minimizing the MI between $\boldsymbol{\hat{x}_{id}}$ and $\boldsymbol{\hat{x}_{a}}%
$. Thus, adversarial training of the embedding $\boldsymbol{\hat{x}_{id}}$ to
minimize the $JSD$ will result in age-invariant embeddings.

To draw samples from the mutual embedding $\mathbb{P}(\boldsymbol{\hat{x}%
_{id}},\boldsymbol{\hat{x}_{a}})$, we follow Chen et al. \cite{infogan} and
combine the embeddings $\boldsymbol{\hat{x}_{id}}$ and $\boldsymbol{\hat
{x}_{a}}$ into a single embedding $\boldsymbol{\hat{x}}$. To create
samples from the statistically independent embeddings $\mathbb{P}%
(\boldsymbol{\hat{x}_{id}})\mathbb{P}(\boldsymbol{\hat{x}_{a}})$, we combine
the embeddings $\boldsymbol{\hat{x}_{id}}$ and $\boldsymbol{\hat{x}_{a}}$ of
\textit{different} images. These are known not to be statistically independent, and we label the result $\boldsymbol{\bar{x}}$. Training a discriminator
using the $JSD$ loss in Eq. \ref{JSD_l} might lead to convergence issues \cite{kodali2017convergence}. For that, we construct a Wasserstein distance discriminator \cite{WDGR}, which utilizes a critic network (MLP) trained with a
Wasserstein distance and a gradient penalty term. The architecture of the
critic layer is detailed in Table 1 of the Appendix, and the training
loss of the discriminator is given by%
\begin{equation}
L_{w}(\mathbb{P}_{M},\mathbb{P}_{I})=\underset{f_{w}}{sup}\,\,\,{E}%
_{\boldsymbol{\hat{x}}\sim\mathbb{P}_{M}}[f_{w}(\boldsymbol{\hat{x}}%
)]-{E}_{x\sim\mathbb{P}_{I}}[f_{w}(\boldsymbol{\hat{x}})]. \label{was}%
\end{equation}
For the sake of brevity, let $\mathbb{P}(\boldsymbol{\hat{x}_{id}%
},\boldsymbol{\hat{x}_{a}})=\mathbb{P}_{M}$ and $\mathbb{P}(\boldsymbol{\hat
{x}_{id}})\mathbb{P}(\boldsymbol{\hat{x}_{a}})=\mathbb{P}_{I}$. The critic network $\boldsymbol{f_{w}}$ is
trained to maximize the Wasserstein distance between $\mathbb{P}_{M}$ and
$\mathbb{P}_{I}$ \cite{WGAN}, and the Wasserstein distance in Eq. \ref{was} is approximated by training the critic network $\boldsymbol{D}$ to maximize
\begin{equation}
L_{w}(\mathbb{P}_{M},\mathbb{P}_{I})=\frac{1}{\left\vert \mathbb{P}%
_{M}\right\vert }\sum_{\boldsymbol{\hat{x}}\in\mathbb{P}_{M}}{f_{w}%
(\boldsymbol{\hat{x}})}-\frac{1}{\left\vert \mathbb{P}_{I}\right\vert }%
\sum_{\boldsymbol{\bar{x}}\in\mathbb{P}_{I}}{{f_{w}(\boldsymbol{\bar{x}}).}}
\label{was1}%
\end{equation}
To maintain the Lipschitz constraint of the Wasserstein distance and improve
the stability of the training process of the critic network, we incorporate a gradient
penalty term \cite{WDGR}. The critic network $\boldsymbol{D}$ and adversarial encoder $\boldsymbol{f_{id}}$ are trained in two stages:
first, we freeze $\boldsymbol{f_{id}}$ and $\boldsymbol{f_{a}}$ and optimize
the critic network $\boldsymbol{D}$ to maximize the Wasserstein distance. We
then freeze the critic network and optimize the $\boldsymbol{f_{id}}$ and
$\boldsymbol{f_{a}}$ networks to minimize the Wasserstein distance and the
$JSD$. The overall optimization is thus given by%
\begin{equation}
L=\underset{\boldsymbol{f_{id}},\boldsymbol{f_{a}}}{min}\,\,\,\,\underset
{\mathbf{D}}{max}(L_{w}-\lambda_{g}L_{grad}), \label{Lw_with_grad}%
\end{equation}
where $L_{grad}$ is the gradient penalty term and we set $\lambda_{g} = 10$ in inline with \cite{WGAN},\cite{WDGR}, where a thorough ablation study of $\lambda_{g}$ was conducted.
\subsection{Face and Age Encoder Architecture}
\label{sec:enc}
The backbone architecture and losses for face recognition and age estimation
were adopted from the ArcFace network \cite{deng2018arcface}.
Thus, we used the Resnet101 backbone for the identity encoder and the Resnet50
network for the age encoder. For the age estimation network $\boldsymbol{g_{a}%
}$, following Rothe et al. \cite{DEX}, we use a single fully connected layer,
which takes as input the normalized age embedding $\boldsymbol{\hat{x}_{a}}$
and is trained using a discrete classification loss $L_{a}$. The identity
prediction network $\boldsymbol{g_{id}}$, also uses a single fully connected
layer, given the identity embedding $\boldsymbol{\hat{x}_{id},}$ and is
trained using the ArcFace loss \cite{deng2018arcface} $L_{id}$.

\subsection{Multitask Training}
\label{sec:MLO}
The embedding encoders and discriminator are jointly trained in an adversarial manner, to compute embeddings that are not discriminative with respect to age, while the discriminator is trained to distinguish between age-invariant and age-dependent embeddings. Through this adversarial training, the MI between the age and identity representations is minimized, and the
resulting identity embeddings are made age-invariant. Our architecture is trained
as a multitask optimization using a min-max term for the discriminator. We
suggest a four-step training:
\begin{enumerate}
\item Forward pass the image $\boldsymbol{x}$ through the encoders to compute $\boldsymbol{\hat{x}_{id}}$ and $\boldsymbol{\hat{x}_{a}}$.
\item Freeze $\boldsymbol{f_{id}}$ and $\boldsymbol{f_{a}}$ and using $\boldsymbol{\hat{x}_{id}}$,$\boldsymbol{\hat{x}_{a}}$, train the discriminator and the critic network $\boldsymbol{D}$ by maximizing the loss
in Eq. \ref{Lw_with_grad}.
\item Freeze $\boldsymbol{f_{a}}$ and forward pass $\boldsymbol{\hat
{x}_{id}}$,$\boldsymbol{\hat{x}_{a}}$ to compute the Wasserstein distance
using the trained discriminator and obtain $L_{w}(\boldsymbol{\hat{x}%
_{id},\hat{x}_{a}})$.
\item To optimize $\boldsymbol{f_{id}}$, first compute $L_{a}$ and $L_{id}$ following Section \ref{sec:enc}. Then, minimize $L_{id}$ and $L_{w}$. To optimize $\boldsymbol{f_{a}}$, minimize $L_{a}$
\end{enumerate}
Thus, the overall training loss is%
\begin{equation}
L=L_{id}(\boldsymbol{\hat{x}_{i\boldsymbol{d}}})+\lambda_{w}L_{w}(\boldsymbol{\hat{x}_{i\boldsymbol{d}},\hat{x}_{a}}%
)+\lambda_{a}L_{a}(\boldsymbol{\hat{x}_{a}}), \label{equ:total loss}%
\end{equation}
where $\lambda_{w}$ and $\lambda_{a}$ are hyperparameters. The ablation of $\lambda_{w}$ is studied in Section \ref{Ablation_Study} where we show $\lambda_{w}=0.1$ to be optimal in terms of verification accuracy. As for $\lambda_{a}$, since we freeze the age embedding network $f_{a}$ while optimizing the Wasserstein distance loss $L_{w}$, the accuracy of our WMI-AI approach is invariant to $\lambda_{a}$, and we set $\lambda_{a} = 1$. The training is summarized in Algorithm 1 in the Appendix.
\subsection{Using a Pretrained Age Estimator}
\label{sec:UGT}
In many situations, the age labels of the face images are unknown. Hence, a
pre-trained age estimation model is utilized to compute the age embeddings
$\boldsymbol{\hat{x}_{a}}$, while the identity embeddings $\boldsymbol{\hat
{x}_{id}}$ are computed the same as in the supervised case in Section
\ref{sec:MLO}. We use two training datasets: the first has both identity and
age labels, while the second lacks age labels. For the second dataset, the
pretrained age estimation network trained on the first dataset is used to
obtain the age embeddings. During training, only the face recognition task is
trained, while $\boldsymbol{\hat{x}_{a}}$ remains fixed for all images. The
Wasserstain discriminator has the same task as in the supervised case,
minimizing the MI between $\boldsymbol{\hat{x}_{id}}$ and the
fixed age embedding $\boldsymbol{\hat{x}_{a}}$. This architecture is shown in
Fig. \ref{fig:full-arc}. The training loss consists of the Wasserstain
discriminator and the identity recognition losses,%
\begin{equation}
L=L_{id}(\boldsymbol{\hat{x}_{id}})+\lambda_{w}L_{w}(\boldsymbol{\hat{x}_{id},\hat{x}_{a}})
\end{equation}
where $\lambda_{w}$ is a hyperparameter (See Section \ref{Ablation_Study}).

\section{Experimental Results}
\label{ExperimentalResults}
\subsection{Training details}
\label{Training_details}
\textbf{Training Parameters:} The proposed scheme was trained using a batch
size of $1,024$ with gradient accumulation to account for the limited GPU memory. We used the SGD optimizer with a momentum value of $0.9$ and a
weight decay of $5\cdot10^{-4}$ for the identity channel, $\boldsymbol{g_{id}%
}$ and $\boldsymbol{f_{id}}$, as well as for the age channel,
$\boldsymbol{g_{a}}$ and $\boldsymbol{f_{a}}$. The initial learning rate was
set at $\lambda=0.1$ for the identity channel and $\lambda=10^{-2}$ for the
age channel, with both values decaying to $0$ during training. The
discriminator was optimized using the RMS prop optimizer with a fixed learning
rate of $\lambda=10^{-4}$, $\alpha$=$0.99$, and $\varepsilon=10^{-8}$, without
momentum. The frequency of discriminator optimization was set at $50$, such that for each backbone optimization step, the discriminator was
optimized by $50$ steps. The hyperparameters were initialized to
$\lambda_{w}=0.1$, $\lambda_{id}=1$, $\lambda_{g}=1$, and $\lambda_{a}=0.25$,
respectively. The training process was conducted using two Nvidia $V100$ GPUs,
each with $32GB$ memory.

\textbf{Training dataset:} As a training dataset we used a clean version of
the widely used MS-Celeb-1M (MS1M) face verification dataset. MS1M consists of
$1M$ identities and more than $10M$ images. Deng et al. \cite{deng2018arcface} suggested using a clean version of MS1M called MS1MV2, which contains
$5.7M$ images and $85K$ identities. This MS1M version has been widely used
in face recognition to achieve state-of-the-art results on multiple
benchmarks. Deng et al. \cite{deng2018arcface} also suggested a clean version of MS1M called MS1MV3, which is an extended version of MS1MV2
that includes $93K$ identities and $5.1M$ images. The images in MS1MV3 were
manually labeled and aligned using $5$ facial landmarks using RetinaFace
\cite{Deng_2020_CVPR}. We trained our architecture using MS1MV2 and MS1MV3,
where all images were aligned, cropped to a
size of $112\times112$, and normalized to a dynamic range of $[-1,1]$. During
training, the images were augmented using random horizontal flipping.

\subsection{Cross-Age Datasets}

\label{Cross_Age_datasets}

\begin{figure*}[tbh]

\centering
\par
\begin{subfigure}{0.19\linewidth} \centering \includegraphics[width=0.45\linewidth]{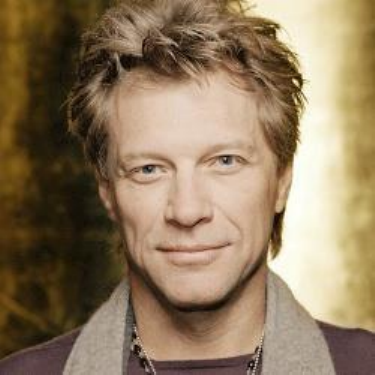} \includegraphics[width=0.45\linewidth]{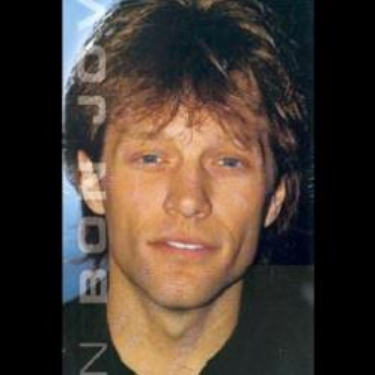} \\ \includegraphics[width=0.45\linewidth]{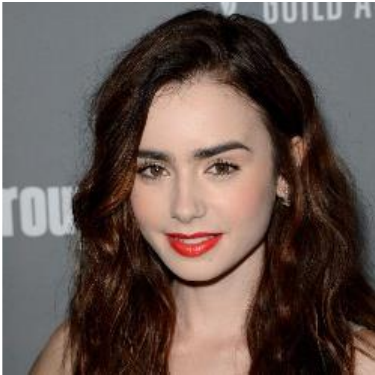} \includegraphics[width=0.45\linewidth]{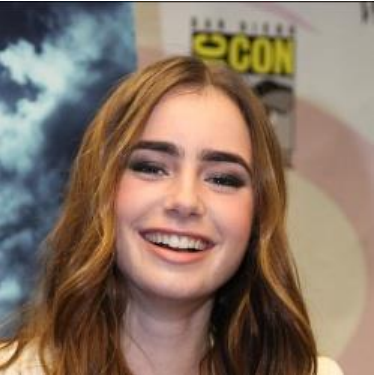} \caption{\scriptsize CACD-VS} \end{subfigure}
\begin{subfigure}{0.19\linewidth} \centering \includegraphics[width=0.45\linewidth]{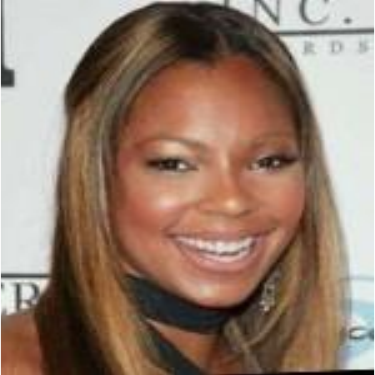} \includegraphics[width=0.45\linewidth]{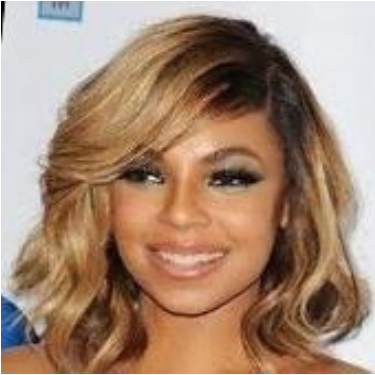} \\ \includegraphics[width=0.45\linewidth]{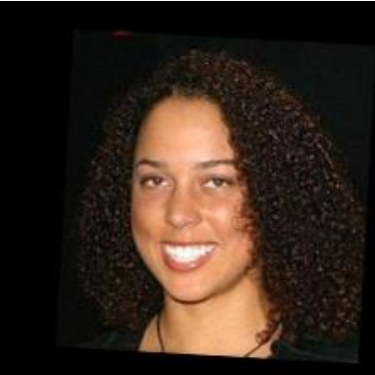} \includegraphics[width=0.45\linewidth]{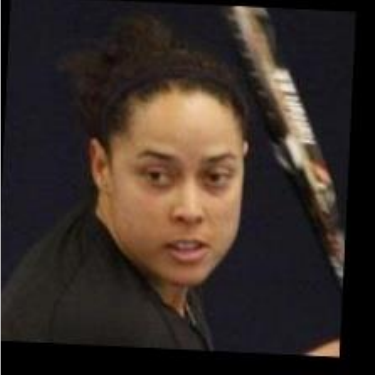} \caption{\scriptsize CALFW} \end{subfigure}
\begin{subfigure}{0.19\linewidth} \centering \includegraphics[width=0.45\linewidth]{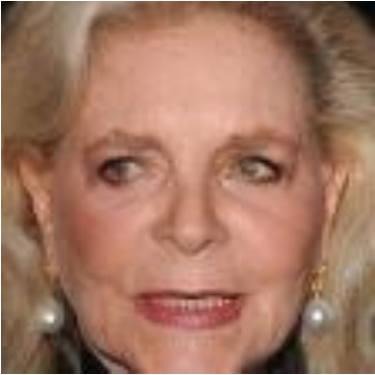} \includegraphics[width=0.45\linewidth]{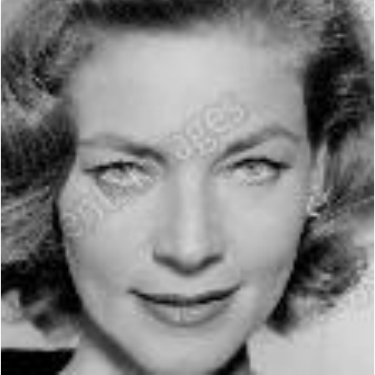} \\ \includegraphics[width=0.45\linewidth]{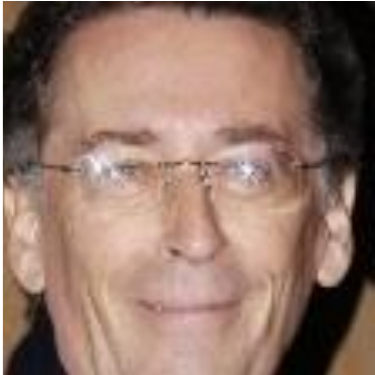} \includegraphics[width=0.45\linewidth]{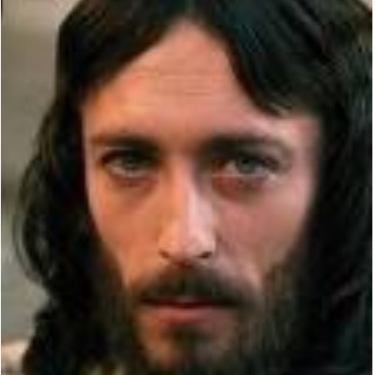} \caption{\scriptsize AGEDB-30} \end{subfigure}
\begin{subfigure}{0.19\linewidth} \centering \includegraphics[width=0.45\linewidth]{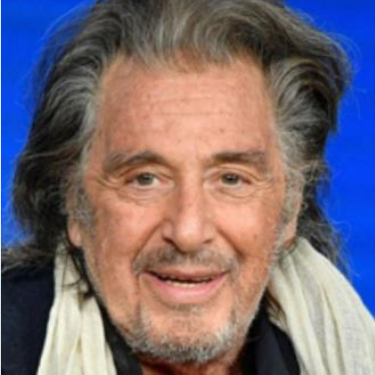} \includegraphics[width=0.45\linewidth]{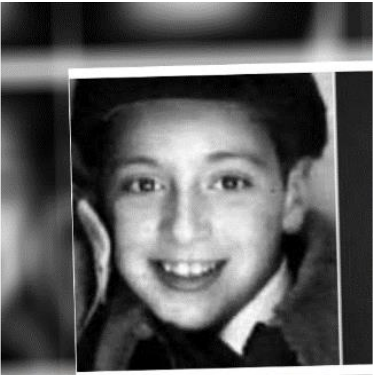} \\ \includegraphics[width=0.45\linewidth]{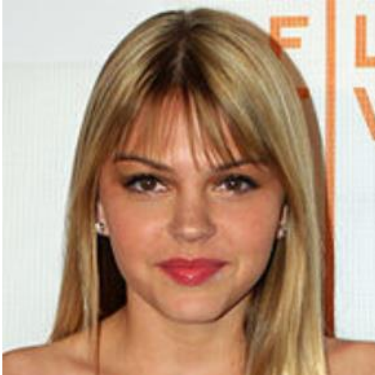} \includegraphics[width=0.45\linewidth]{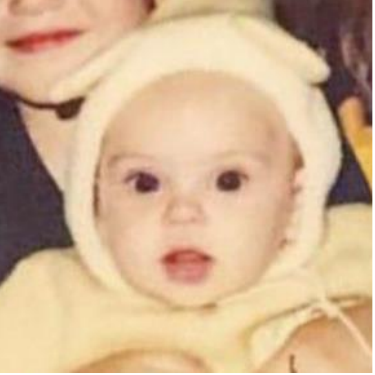} \caption{\scriptsize ECAF(Adult,Child)} \end{subfigure}
\begin{subfigure}{0.19\linewidth} \centering \includegraphics[width=0.45\linewidth]{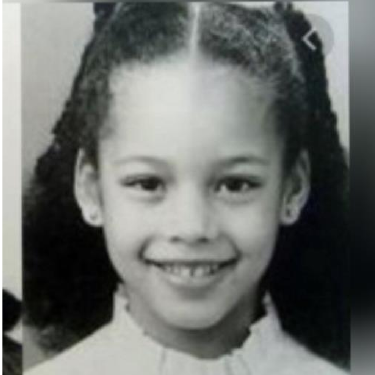} \includegraphics[width=0.45\linewidth]{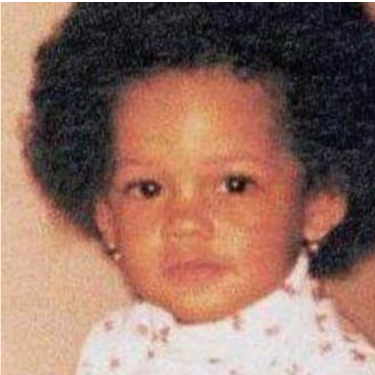} \\ \includegraphics[width=0.45\linewidth]{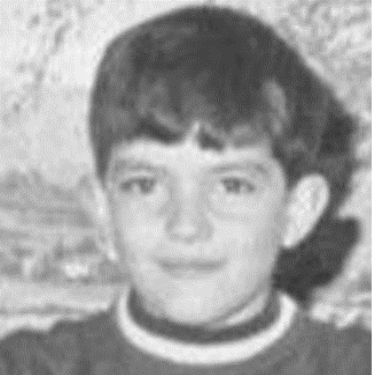} \includegraphics[width=0.45\linewidth]{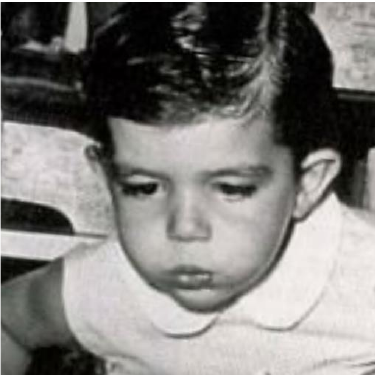} \caption{\scriptsize ECAF(Child,Child)} \end{subfigure}
\caption{Qualitative face verification results using different image datasets.
The first row shows true positive verification results, while, the second row
shows false negative verification results.}%
\label{fig:true_false}%
\end{figure*}

Our proposed method was evaluated on various cross-age datasets, including
CALFW \cite{CALFW}, AgeDB-30 \cite{agedb30}, CACD-VS \cite{cacdvs}, and ECAF
\cite{learning}. Qualitative verification results from each dataset are presented in
Fig. \ref{fig:true_false}. These datasets contain images with a significant age
gaps, with the ECAF (adult, child) dataset having an average age gap of $41$
years. The images are captured \textquotedblleft in the wild\textquotedblright%
\ in different scenarios, featuring diverse poses, lighting conditions, and
facial expressions, some available in either black-and-white or color.

\textbf{CALFW:} The Cross-Age Labeled Faces in the Wild (CALFW) dataset
\cite{CALFW} is a subset of the Labeled Faces in the Wild (LFW) dataset
\cite{LFWTech}. The CALFW dataset consists of $13,216$ images of $574$
individuals, with ages ranging from $0$ to $100$ years. The dataset has
$3,000$ positive pairs of images with large age gaps, the evaluation protocol
uses $10$ folds of verification, each fold consisting of $600$ positive and
negative pairs. We follow this protocol to evaluate our method, and the
results are summarized in Table \ref{table:CALFW}. All of the verification results in Table \ref{table:CALFW} are cited from the corresponding papers, except for the ArcFace \cite{deng2020subcenter} results using the MS1MV3 dataset which was computed with the publicly available ArcFace code\footnote{\url{https://github.com/deepinsight/insightface}}.
Two versions of our proposed WMI-AI scheme were trained using either the MS1MV2 or MS1MV3 datasets
separately. In both cases, we observed an accuracy improvement of 0.06\% and
0.02\%, respectively, compared to the SOTA. In
particular, when training with the MS1MV3 dataset, the verification accuracy
improved consistently between different methods.

\def\hfillx{\hspace*{-\textwidth}\hfill}
\begin{table*}[tbh]
\begin{minipage}{0.45\textwidth}
\vspace{-10pt}
\centering
\caption{Face verification results evaluated using the CALFW dataset \cite{CALFW}. The
leading results are highlighted in \textbf{bold}.}%
\label{table:CALFW}%
\centering
\begin{tabular}
[c]{lcc}%
\toprule Method & Dataset & Accuracy[\%]\\
\midrule MTLFace\cite{learning} & \multirow{7}{*}{\scriptsize MS1MV2} & 95.62\\
Arc-DiscFace\cite{discface} &  & 96.15\\
ElasticFace\cite{elastic} &  & 96.18\\
ArcFace\cite{deng2020subcenter} &  & 96.18\\
CosFace\cite{deng2018arcface} &  & 96.20\\
CurricularFace\cite{huang2020curricularface} &  & 96.20\\
\textbf{WMI-AI (ours)} &  & \textbf{96.26}\\\midrule
Implicit\cite{9682736} & \multirow{4}{*}{\scriptsize MS1MV3} & 95.82\\
DPNN\cite{Chrysos_2021} &  & 96.23\\
ArcFace\cite{deng2020subcenter} &  & 96.26\\
\textbf{WMI-AI (ours)} &  & \textbf{96.28}\\
\bottomrule &  &
\end{tabular}
        \end{minipage}%
        \hfillx
        \begin{minipage}{0.45\textwidth}
            \centering
        \caption{Face verification results evaluated using the AGEDB-30 dataset \cite{agedb30}. The
leading results are highlighted in \textbf{bold}.}%
\label{table:AGEDB}
\centering
\begin{tabular}
[c]{lcc}%
\toprule Method & Dataset & Accuracy[\%]\\
\midrule MTLFACE\cite{agedb30} & \multirow{8}{*}{\scriptsize MS1MV2} & 96.23\\
CosFace\cite{deng2018arcface} &  & 98.07\\
MagFace\cite{meng2021magface} &  & 98.17\\
ArcFace\cite{deng2020subcenter} &  & 98.2\\
CurricularFace\cite{huang2020curricularface} &  & 98.32\\
Arc-DiscFace\cite{discface} &  & 98.35\\
ElasticFace\cite{elastic} &  & 98.35\\
\textbf{WMI-AI (ours)} &  & \textbf{98.37}\\\midrule
Implicit\cite{9682736} & \multirow{4}{*}{\scriptsize MS1MV3} & 95.82\\
DPNN\cite{Chrysos_2021} &  & 98.47\\
ArcFace\cite{deng2020subcenter} &  & 98.55\\
\textbf{WMI-AI (ours)} &  & \textbf{98.57}\\
\bottomrule &  &
\end{tabular}
\end{minipage}%
\vspace{-10pt}
\end{table*}

\textbf{AgeDB-30:} The AgeDB dataset \cite{agedb30} contains $16,488$ images
of $568$ identities that were manually annotated and labeled. Four protocols
for age-variant face verification are used with this dataset, where each
protocol is subject to a different age gap. AgeDB-30, a subset of the AgeDB
dataset, is the most challenging protocol with a 30-year age gap between the
corresponding image pairs. In total, AgeDB-30 has $6K$ pairs divided into
$10$-fold verification sets. The evaluation results are summarized in Table
\ref{table:AGEDB}, where all of the results are cited from the corresponding papers. The results show that our approach achieves an incremental
improvement compared to the state-of-the-art results for both training
datasets, as in the case of the CALFW data set, the training results on the
cleaned MS1MV3 dataset outperform those of the MS1MV2
dataset.

\textbf{CACD-VS:} Cross-Age Celebrity Dataset - Verification Subset
\cite{cacdvs} is a subset of the larger CACD dataset \cite{chen14cross},
consisting of $163,446$ images of $2,000$ celebrities. The CACD-VS subset is
manually sampled and annotated and contains $4K$ pairs of positive and
negative face images. The face verification protocol is similar to LFW and the
results are summarized in Table \ref{table:CACD-VS}. Except for the verification outcomes of ArcFace \cite{deng2020subcenter} on the MS1MV3 dataset, which were computed using the publicly accessible ArcFace code, the verification results were obtained from their respective papers. The results show
that the proposed WMI-AI achieves $99.57\%$ accuracy, outperforming the previous state-of-the-art methods such as MTLFACE
\cite{learning} and ArcFace \cite{deng2020subcenter}. The WMI-AI accuracy is
consistently high across both training datasets, confirming the effectiveness of our approach in
minimizing the intraclass age-related variations.
\def\hfillx{\hspace*{-\textwidth}\hfill}
\begin{table*}[tb]
\begin{minipage}{0.45\textwidth}
\vspace{-25pt}
\centering
\caption{Face verification results evaluated using the CACD-VS dataset \cite{cacdvs}. The
leading results are highlighted in \textbf{bold}.}%
\label{table:CACD-VS}
\centering
\begin{tabular}
[c]{lcc}%
\toprule Method & Dataset & Accuracy [\%]\\
\midrule MTLFACE\cite{learning} & \multirow{3}{*}{\footnotesize MS1MV2} & 99.55\\
ArcFace\cite{deng2020subcenter} &  & 99.55\\
\textbf{WMI-AI (ours)} &  & \textbf{99.57}\\\midrule
ArcFace\cite{deng2020subcenter} & \multirow{3}{*}{\footnotesize MS1MV3} & 99.55\\
Implicit\cite{9682736} &  & \textbf{99.57}\\
\textbf{WMI-AI (ours)} &  & \textbf{99.57}\\
\bottomrule &  &
\end{tabular}
        \end{minipage}%
        \hfillx
        \begin{minipage}{0.45\textwidth}
            \centering
        \caption{Face verification results evaluated using the ECAF dataset \cite{learning}. The
leading results are highlighted in \textbf{bold}.}%
\label{table:ECAF}
\centering
\begin{tabular}
[c]{lcc}%
\toprule \multirow{2}{*}{Method} & (Adult, & (Child\\
& ,Child) & ,Child)\\
\midrule Human, Average\cite{learning} & 73.34 & 68.62\\
Human, Voting\cite{learning} & 85.95 & 78.75\\
Softmax\cite{learning} & 85.03 & 88.25\\
CosFace\cite{deng2018arcface} & 85.72 & 90.75\\
ArcFace\cite{deng2020subcenter} & 86.52 & 90.65\\
CurricularFace\cite{huang2020curricularface} & 84.78 & 90.80\\
MTLFace\cite{learning} & 87.55 & 91.20\\
\textbf{WMI-AI (ours)} & \textbf{90.73} & \textbf{94.44}\\
\bottomrule &  &
\end{tabular}
\end{minipage}%
\vspace{-10pt}
\end{table*}

\textbf{ECAF Dataset:} The ECAF (Evaluation of Cross-Age Face) dataset
\cite{learning} is a recent face recognition dataset introduced by Haung et
al. \cite{learning} that comprises a diverse range of ages, from children to
adults. It consists of two subsets, ECAF-AdultChild and ECAF-ChildChild, of
adult-child and child-child face image pairs. The ECAF is the largest
cross-age dataset with an average age difference of $41$ years, consisting of
$5,265$ images from $613$ identities, resulting in $6K$ (Adult, Child) and
$4K$ (Child, Child) image pairs. In this study, we adopt the evaluation protocol utilized in LFW and assess the performance of our network using the MS1MV2 dataset since there are no previous results reported for the MS1MV3 dataset. Our evaluation results are reported in Table \ref{table:ECAF}, where the results are cited from the respective papers.
The proposed WMI-AI approach shows significant improvements in the
verification results on the challenging ECAF dataset, with an average accuracy
increase of $3.18$\% for the (Adult, Child) pairs and $3.24$\% for the (Child,
Child) pairs compared to the current state-of-the-art result. These results
exemplify the effectiveness of our approach in handling large age differences
between images, as well as achieving age-invariant representations, thus
improving the face verification accuracy.
\begin{table*}[tbh]
\caption{Ablation study of the verification accuracy results [\%] on the
cross-age datasets with respect to $\lambda_{w}.$}%
\label{table:lamda-w}%
\centering
\begin{tabular}
[c]{llccccc}%
\toprule & $\lambda_{w}$ & CALFW & AGEDB30 & CACD-VS & ECAF(A,C) & ECAF(C,C)\\
\midrule\multirow{4}{*}{{\rotatebox[origin=c]{90}{MS1MV2}}} & $0$ & 96.18 &
98.20 & 95.55 & 90.43 & 94.00\\
& $0.1$ & \textbf{96.26} & \textbf{98.37} & 99.57 & \textbf{90.73} &
\textbf{94.40}\\
& $1.0$ & 96.21 & 98.28 & 99.57 & 90.41 & 94.20\\
& $2.0$ & 96.15 & 98.20 & 99.55 & 89.80 & 94.10\\
\midrule\multirow{4}{*}{{\rotatebox[origin=c]{90}{MS1MV3}}} & $0$ & 96.26 &
98.55 & 99.55 & 92.35 & 95.05\\
& $0.1$ & \textbf{96.28} & \textbf{98.57} & 99.57 & \textbf{92.70} &
\textbf{95.20}\\
& $1.0$ & 96.25 & 98.47 & 99.55 & 91.60 & 94.70\\
& $2.0$ & 96.18 & 98.20 & 99.50 & 91.55 & 94.50\\
\bottomrule &  &  &  &  &  &
\end{tabular}
\vspace{-5pt}
\end{table*}
\subsection{Ablation Study}
\label{Ablation_Study}
We examined the impact of the discriminator architectures, as well as the effect of the
weight of the discriminator loss $\lambda_{w},$ in Eq. \ref{equ:total loss},
with respect to the other training losses. We tested the performance of our
approach with different values of $\lambda_{w}$ in Table \ref{table:lamda-w},
including $\lambda_{w}=0$ that corresponds to training our WMI-AI approach
without disentanglement. $\lambda_{w}=0.1$ achieves the highest verification accuracy, showing a \textquotedblleft sweetspot\textquotedblright%
\ effect, where using a higher or lower $\lambda_{w}$ reduces the accuracy.

\begin{figure}[tb]
\centering
\begin{subfigure}{0.40\textwidth}
\centering
\includegraphics[width=\linewidth]{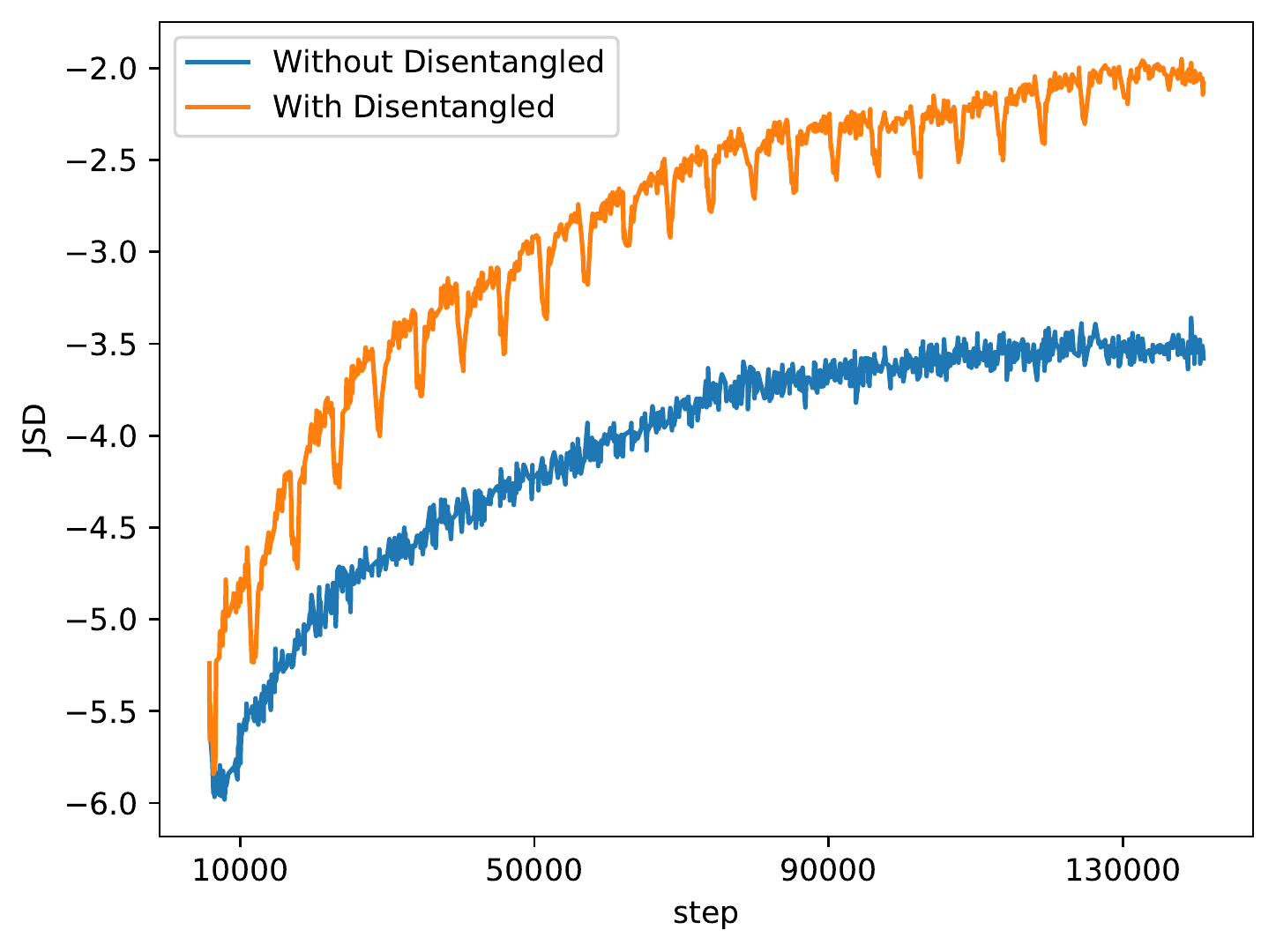}
\caption{MS1MV2}
\label{fig:ms1mv2_compare}
\end{subfigure}
\hfill\begin{subfigure}{0.40\textwidth}
\centering
\includegraphics[width=\linewidth]{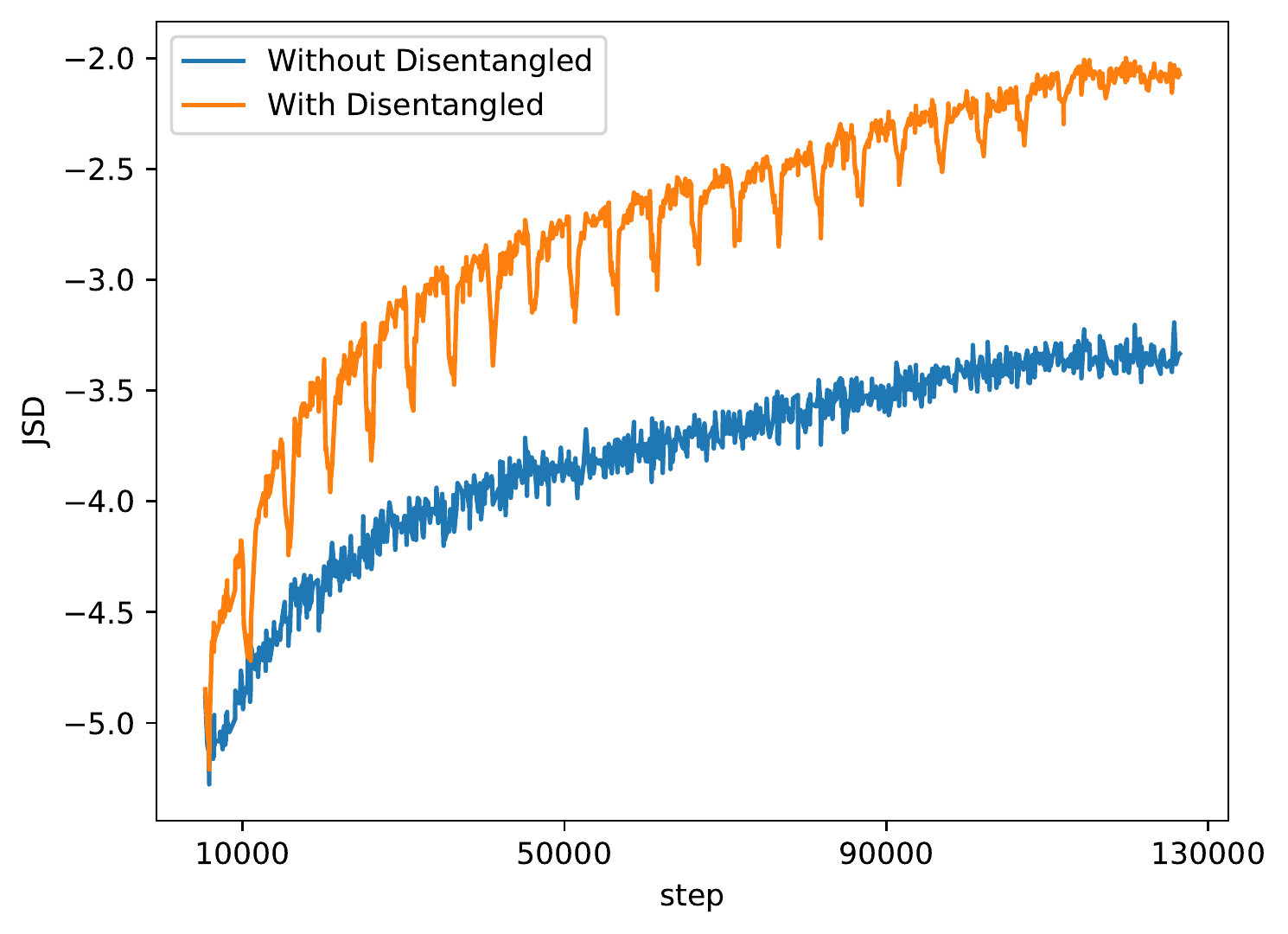}
\caption{MS1MV3}
\label{fig:ms1mv3_compare}
\end{subfigure}
\caption{The Jensen-Shannon divergence (JSD) with ($\lambda_{w}=0.1$) and
without ($\lambda_{w}=0$) disentanglement when trained using the MS1MV2 and
MS1MV3 datasets. We report the JSD with respect to the optimization steps of
our scheme.}%
\label{fig:distance}%
\vspace{-10pt}
\end{figure}

We also evaluated the use of the proposed disentanglement on the
Jensen-Shannon divergence (JSD), which measures the MI between
age and identity embeddings. The results for minimizing the JSD when training
the proposed network with ($\lambda_{w}=0.1$) and without ($\lambda_{w}=0$)
disentanglement using the MS1MV2 and MS1MV3 datasets are shown in Fig.
\ref{fig:distance}. The proposed scheme better minimized the MI in both cases. It is interesting to note that applying our network
without the disentanglement ($\lambda_{w}=0$) also reduces the JSD, as the
identity and age encoders $\boldsymbol{f_{id}}$ and $\boldsymbol{f_{a}}$,
respectively, compute identity-specific and age-specific embeddings, thus increasing
the JSD between them. However, applying the proposed
disentanglement is shown to better disentangle these
embeddings statistically.

An Ablation of the WMI backbone and critic network is reported in Table 2 of the Appendix.
\section{Conclusions and Future Work}
\label{Conclusion}
This paper proposes a new approach to age-invariant face embedding based on the Wasserstein distance. Our approach disentangles age and identity
embeddings of face images, enabling improved face verification performance in age-variant datasets. We apply multitask training with a Wasserstein distance discriminator that minimizes the MI between age and identity embeddings by minimizing the Jensen-Shannon divergence. Our experimental results demonstrate the
effectiveness of the proposed method, with superior face verification accuracy compared to state-of-the-art methods on multiple contemporary age-variant face
datasets. Thus, we believe that the proposed WMI-AI method offers a promising solution to face verification in datasets with significant age-varying images of the same subjects. Future work could study the applicability of our approach to domains beyond face verification, such as kinship verification, where age variance plays a significant role in identifying family-related face image pairs.
\bibliographystyle{plain}
\bibliography{WMI-AI}

\begin{thebibliography}{10}

\bibitem{WGAN}
Martin Arjovsky, Soumith Chintala, and L{\'e}on Bottou.
\newblock Wasserstein generative adversarial networks.
\newblock In {\em {Proceedings of the International Conference on Machine
  Learning ({ICML})}}, pages 214--223. PMLR, 2017.

\bibitem{medicine}
Joe Bathelt, P.~Cédric Koolschijn, and Hilde~M Geurts.
\newblock Age-variant and age-invariant features of functional brain
  organization in middle-aged and older autistic adults.
\newblock {\em {Aging Ment Health}}, 11(1), 2020.

\bibitem{belghazi2018mutual}
Mohamed~Ishmael Belghazi, Aristide Baratin, Sai Rajeshwar, Sherjil Ozair,
  Yoshua Bengio, Aaron Courville, and Devon Hjelm.
\newblock Mutual information neural estimation.
\newblock In {\em {Proceedings of the International Conference on Machine
  Learning ({ICML})}}, pages 531--540. PMLR, 2018.

\bibitem{elastic}
Fadi Boutros, Naser Damer, Florian Kirchbuchner, and Arjan Kuijper.
\newblock {ElasticFace:} elastic margin loss for deep face recognition.
\newblock In {\em {Proceedings of the {IEEE} Conference on Computer Vision and
  Pattern Recognition Workshops ({CVPRW})}}, pages 1577--1586, 2022.

\bibitem{lostchild}
Praveen~Kumar Chandaliya and Neeta Nain.
\newblock Childgan: Face aging and rejuvenation to find missing children.
\newblock {\em {Pattern Recognition}}, 129:108761, 2022.

\bibitem{cacdvs}
Bor-Chun Chen, Chu-Song Chen, and Winston~H. Hsu.
\newblock Cross-age reference coding for age-invariant face recognition and
  retrieval.
\newblock In {\em {Proceedings of the European Conference on Computer Vision
  ({ECCV})}}, 2014.

\bibitem{chen14cross}
Bor-Chun Chen, Chu-Song Chen, and Winston~H. Hsu.
\newblock Cross-age reference coding for age-invariant face recognition and
  retrieval.
\newblock In {\em Proceedings of the European Conference on Computer Vision
  ({ECCV})}, 2014.

\bibitem{infogan}
Xi~Chen, Yan Duan, Rein Houthooft, John Schulman, Ilya Sutskever, and Pieter
  Abbeel.
\newblock {InfoGAN:} interpretable representation learning by information
  maximizing generative adversarial nets.
\newblock In {\em {Neural Information Processing Systems ({NeurIPS})}},
  NIPS'16, page 2180–2188, 2016.

\bibitem{cheng2020club}
Pengyu Cheng, Weituo Hao, Shuyang Dai, Jiachang Liu, Zhe Gan, and Lawrence
  Carin.
\newblock Club: A contrastive log-ratio upper bound of mutual information.
\newblock In {\em {Proceedings of the International Conference on Machine
  Learning ({ICML})}}, pages 1779--1788. PMLR, 2020.

\bibitem{Chrysos_2021}
Grigorios~G. Chrysos, Stylianos Moschoglou, Giorgos Bouritsas, Jiankang Deng,
  Yannis Panagakis, and Stefanos~P Zafeiriou.
\newblock Deep polynomial neural networks.
\newblock {\em {{IEEE} Transactions on Pattern Analysis and Machine
  Intelligence ({PAMI})}}, pages 1--1, 2021.

\bibitem{9411913}
Debayan Deb, Divyansh Aggarwal, and Anil~K. Jain.
\newblock Identifying missing children: Face age-progression via deep feature
  aging.
\newblock In {\em {Proceedings of the International Conference on Pattern
  Recognition ({ICPR})}}, pages 10540--10547, 2021.

\bibitem{deng2020subcenter}
Jiankang Deng, Jia Guo, Tongliang Liu, Mingming Gong, and Stefanos Zafeiriou.
\newblock {Sub-center ArcFace:} boosting face recognition by large-scale noisy
  web faces.
\newblock In {\em {Proceedings of the European Conference on Computer Vision
  ({ECCV})}}, 2020.

\bibitem{deng2018arcface}
Jiankang Deng, Jia Guo, Xue Niannan, and Stefanos Zafeiriou.
\newblock {ArcFace:} additive angular margin loss for deep face recognition.
\newblock In {\em {Proceedings of the {IEEE} Conference on Computer Vision and
  Pattern Recognition ({CVPR})}}, 2019.

\bibitem{Deng_2020_CVPR}
Jiankang Deng, Jia Guo, Evangelos Ververas, Irene Kotsia, and Stefanos
  Zafeiriou.
\newblock {RetinaFace:} single-shot multi-level face localisation in the wild.
\newblock In {\em {Proceedings of the {IEEE} Conference on Computer Vision and
  Pattern Recognition ({CVPR})}}, June 2020.

\bibitem{scene}
Konstantinos Drossos, Paul Magron, and Tuomas Virtanen.
\newblock Unsupervised adversarial domain adaptation based on the wasserstein
  distance for acoustic scene classification.
\newblock In {\em {Proceedings of the {IEEE} Workshop on Applications of Signal
  Processing to Audio and Acoustics}}, pages 259--263. IEEE, 2019.

\bibitem{GEPSHTEIN2019368}
Shai Gepshtein and Yosi Keller.
\newblock Iterative spectral independent component analysis.
\newblock {\em Signal Processing}, 155:368--376, 2019.

\bibitem{Gong}
Dihong Gong, Zhifeng Li, Dahua Lin, Jianzhuang Liu, and Xiaoou Tang.
\newblock Hidden factor analysis for age invariant face recognition.
\newblock In {\em {Proceedings of the {IEEE} Conference on Computer Vision and
  Pattern Recognition ({CVPR})}}, pages 2872--2879, 2013.

\bibitem{Gong_new}
Dihong Gong, Zhifeng Li, Dacheng Tao, Jianzhuang Liu, and Xuelong Li.
\newblock A maximum entropy feature descriptor for age invariant face
  recognition.
\newblock In {\em {Proceedings of the {IEEE} Conference on Computer Vision and
  Pattern Recognition ({CVPR})}}, pages 5289--5297, 2015.

\bibitem{Hou}
Xuege Hou, Yali Li, and Shengjin Wang.
\newblock Disentangled representation for age-invariant face recognition: A
  mutual information minimization perspective.
\newblock In {\em {Proceedings of the {IEEE} Conference on Computer Vision and
  Pattern Recognition ({CVPR})}}, pages 3692--3701, 2021.

\bibitem{LFWTech}
Gary~B. Huang, Manu Ramesh, Tamara Berg, and Erik Learned-Miller.
\newblock Labeled faces in the wild: A database for studying face recognition
  in unconstrained environments.
\newblock Technical Report 07-49, University of Massachusetts, Amherst, October
  2007.

\bibitem{huang2020curricularface}
Yuge Huang, Yuhan Wang, Ying Tai, Xiaoming Liu, Pengcheng Shen, Shaoxin Li, and
  Feiyue~Huang Jilin~Li.
\newblock {CurricularFace:} adaptive curriculum learning loss for deep face
  recognition.
\newblock In {\em {Proceedings of the {IEEE} Conference on Computer Vision and
  Pattern Recognition ({CVPR})}}, June 2020.

\bibitem{learning}
Zhizhong Huang, Junping Zhang, and Hongming Shan.
\newblock When age-invariant face recognition meets face age synthesis: A
  multi-task learning framework.
\newblock In {\em {Proceedings of the {IEEE} Conference on Computer Vision and
  Pattern Recognition ({CVPR})}}, 2021.

\bibitem{fastica}
A.~Hyvarinen.
\newblock Fast and robust fixed-point algorithms for independent component
  analysis.
\newblock {\em IEEE Transactions on Neural Networks}, 10(3):626--634, 1999.

\bibitem{discface}
Insoo Kim, Seungju Han, Seong-Jin Park, Ji-won Baek, Jinwoo Shin, Jae-Joon Han,
  and Changkyu Choi.
\newblock {DiscFace:} minimum discrepancy learning for deep face recognition.
\newblock In Hiroshi Ishikawa, Cheng-Lin Liu, Tomas Pajdla, and Jianbo Shi,
  editors, {\em {Proceedings of the Asian Conference on Computer Vision
  ({ACCV})}}, pages 358--374, 2021.

\bibitem{kodali2017convergence}
Naveen Kodali, Jacob Abernethy, James Hays, and Zsolt Kira.
\newblock On convergence and stability of {GANs}.
\newblock {\em arXiv preprint arXiv:1705.07215}, 2017.

\bibitem{Lee2021ImprovingFR}
Jungsoo Lee, Jooyeol Yun, Sunghyun Park, Yonggyu Kim, and Jaegul Choo.
\newblock Improving face recognition with large age gaps by learning to
  distinguish children.
\newblock In {\em {Proceedings of the British Machine Vision Conference
  ({BMVC})}}, 2021.

\bibitem{Ling}
Haibin Ling, Stefano Soatto, Narayanan Ramanathan, and David~W. Jacobs.
\newblock Face verification across age progression using discriminative
  methods.
\newblock {\em {{IEEE} Transactions on Information Forensics and Security}},
  5(1):82--91, 2010.

\bibitem{kinship}
Fan Liu, Zewen Li, Wenjie Yang, and Feng Xu.
\newblock Age-invariant adversarial feature learning for kinship verification.
\newblock {\em {Mathematics}}, 10(3), 2022.

\bibitem{meng2021magface}
Qiang Meng, Shichao Zhao, Zhida Huang, and Feng Zhou.
\newblock {MagFace}: A universal representation for face recognition and
  quality assessment.
\newblock In {\em {Proceedings of the {IEEE} Conference on Computer Vision and
  Pattern Recognition ({CVPR})}}, 2021.

\bibitem{medicine2}
Dario Moreno-Agostino, Alejandro de~la Torre-Luque, Leandro da~Silva-Sauer,
  Bruce~W Smith, and Bernardino Fernández-Calvo.
\newblock The age-invariant role of resilience resources in emotional
  symptomatology.
\newblock {\em {Molecular Autism}}, 26(1), 2022.

\bibitem{agedb30}
Stylianos Moschoglou, Athanasios Papaioannou, Christos Sagonas, Jiankang Deng,
  Irene Kotsia, and Stefanos Zafeiriou.
\newblock Agedb: the first manually collected, in-the-wild age database.
\newblock In {\em {Proceedings of the {IEEE} Conference on Computer Vision and
  Pattern Recognition Workshops ({CVPRW})}}, volume~2, page~5, 2017.

\bibitem{DEEPID}
W.~Ouyang, X.~Wang, X.~Zeng, Shi Qiu, P.~Luo, Y.~Tian, H.~Li, Shuo Yang, Zhe
  Wang, Chen-Change Loy, and X.~Tang.
\newblock {DeepID-Net:} deformable deep convolutional neural networks for
  object detection.
\newblock In {\em {Proceedings of the {IEEE} Conference on Computer Vision and
  Pattern Recognition ({CVPR})}}, pages 2403--2412, jun 2015.

\bibitem{Park}
Unsang Park, Yiying Tong, and Anil~K. Jain.
\newblock Age-invariant face recognition.
\newblock {\em {{IEEE} Transactions on Pattern Analysis and Machine
  Intelligence ({PAMI})}}, 32(5):947--954, 2010.

\bibitem{Ramanathan}
Narayanan Ramanathan and Rama Chellappa.
\newblock Modeling shape and textural variations in aging faces.
\newblock In {\em {Proceedings of the {IEEE} International Conference on
  Automatic Face and Gesture Recognition}}, pages 1--8, 2008.

\bibitem{vgg_triplet_loss_ref}
Joseph~P Robinson, Ming Shao, Yue Wu, and Yun Fu.
\newblock Families in the wild ({FIW}): Large-scale kinship image database and
  benchmarks.
\newblock In {\em {Proceedings of the {ACM} International Conference on
  Multimedia}}, pages 242--246. ACM, 2016.

\bibitem{DEX}
Rasmus Rothe, Radu Timofte, and Luc Van~Gool.
\newblock {DEX:} deep expectation of apparent age from a single image.
\newblock In {\em {Proceedings of the International Conference on Computer
  Vision Workshops ({ICCVW})}}, pages 252--257, 2015.

\bibitem{demographic}
Muhammad Sajid, Tamoor Shafique, Sohaib Manzoor, Faisal Iqbal, Hassan Talal,
  Usama Samad~Qureshi, and Imran Riaz.
\newblock Demographic-assisted age-invariant face recognition and retrieval.
\newblock {\em {Symmetry}}, 10(5), 2018.

\bibitem{WDGR}
Jian Shen, Yanru Qu, Weinan Zhang, and Yong Yu.
\newblock Wasserstein distance guided representation learning for domain
  adaptation.
\newblock In {\em {Proceedings of the {AAAI} Conference on Artificial
  Intelligence ({AAAI})}}, AAAI'18/IAAI'18/EAAI'18, 2018.

\bibitem{shwartz2017opening}
Ravid Shwartz{-}Ziv and Naftali Tishby.
\newblock Opening the black box of deep neural networks via information.
\newblock {\em CoRR}, abs/1703.00810, 2017.

\bibitem{rep2}
Yu~Su, Shiguang Shan, Xilin Chen, and Wen Gao.
\newblock Hierarchical ensemble of global and local classifiers for face
  recognition.
\newblock {\em {{IEEE} Transactions on Image Processing}}, 18(8):1885--1896,
  2009.

\bibitem{rep1}
S.~N. Sujay, H.~S.~Manjunatha Reddy, and J.~Ravi.
\newblock Face recognition using extended lbp features and multilevel svm
  classifier.
\newblock In {\em {Proceedings of the {IEEE} International Conference on
  Electrical, Electronics, Communication, Computer, and Optimization
  Techniques}}, pages 1--4, 2017.

\bibitem{deepface}
Yaniv Taigman, Ming Yang, Marc'Aurelio Ranzato, and Lior Wolf.
\newblock {DeepFace:} closing the gap to human-level performance in face
  verification.
\newblock In {\em {Proceedings of the {IEEE} Conference on Computer Vision and
  Pattern Recognition ({CVPR})}}, pages 1701--1708, 2014.

\bibitem{7133169}
Naftali Tishby and Noga Zaslavsky.
\newblock Deep learning and the information bottleneck principle.
\newblock In {\em {Proceedings of the {IEEE} Information Theory Workshop}},
  pages 1--5, 2015.

\bibitem{8954207}
Hao Wang, Dihong Gong, Zhifeng Li, and Wei Liu.
\newblock Decorrelated adversarial learning for age-invariant face recognition.
\newblock In {\em {Proceedings of the {IEEE} Conference on Computer Vision and
  Pattern Recognition ({CVPR})}}, pages 3522--3531, 2019.

\bibitem{Cosface}
Hao Wang, Yitong Wang, Zheng Zhou, Xing Ji, Dihong Gong, Jingchao Zhou, Zhifeng
  Li, and Wei Liu.
\newblock {CosFace}: Large margin cosine loss for deep face recognition.
\newblock In {\em {Proceedings of the {IEEE} Conference on Computer Vision and
  Pattern Recognition ({CVPR})}}, pages 5265--5274, 2018.

\bibitem{kinship2}
Shuyang Wang, Zhengming Ding, and Yun Fu.
\newblock Cross-generation kinship verification with sparse discriminative
  metric.
\newblock {\em {{IEEE} Transactions on Pattern Analysis and Machine
  Intelligence ({PAMI})}}, 41(11):2783--2790, 2019.

\bibitem{Wang1}
Yitong Wang, Dihong Gong, Zheng Zhou, Xing Ji, Hao Wang, Zhifeng Li, Wei Liu,
  and Tong Zhang.
\newblock Orthogonal deep features decomposition for age-invariant face
  recognition.
\newblock In Vittorio Ferrari, Martial Hebert, ristian Sminchisescu, and Yair
  Weiss, editors, {\em {Proceedings of the European Conference on Computer
  Vision ({ECCV})}}, pages 764--779, 2008.

\bibitem{7780898}
Yandong Wen, Zhifeng Li, and Yu~Qiao.
\newblock Latent factor guided convolutional neural networks for age-invariant
  face recognition.
\newblock In {\em {Proceedings of the {IEEE} Conference on Computer Vision and
  Pattern Recognition ({CVPR})}}, pages 4893--4901, 2016.

\bibitem{multi-source}
Hanrui Wu, Yuguang Yan, Michael~K Ng, and Qingyao Wu.
\newblock Domain-attention conditional wasserstein distance for multi-source
  domain adaptation.
\newblock {\em {{ACM} Transactions on Intelligent Systems and Technology}},
  11(4):1--19, 2020.

\bibitem{9682736}
Jiu-Cheng Xie, Chi-Man Pun, and Kin-Man Lam.
\newblock Implicit and explicit feature purification for age-invariant facial
  representation learning.
\newblock {\em {{IEEE} Transactions on Information Forensics and Security}},
  17:399--412, 2022.

\bibitem{object}
Pengcheng Xu, Prudhvi Gurram, Gene~T. Whipps, and Rama Chellappa.
\newblock Wasserstein distance based domain adaptation for object detection.
\newblock {\em CoRR}, abs/1909.08675, 2019.

\bibitem{9146699}
Jian Zhao, Shuicheng Yan, and Jiashi Feng.
\newblock Towards age-invariant face recognition.
\newblock {\em {{IEEE} Transactions on Pattern Analysis and Machine
  Intelligence ({PAMI})}}, 44(1):474--487, 2022.

\bibitem{CALFW}
Tianyue Zheng, Weihong Deng, and Jiani Hu.
\newblock Cross-age {LFW:} {A} database for studying cross-age face recognition
  in unconstrained environments.
\newblock {\em CoRR}, abs/1708.08197, 2017.

\end{thebibliography}


\includepdf[pages=-]{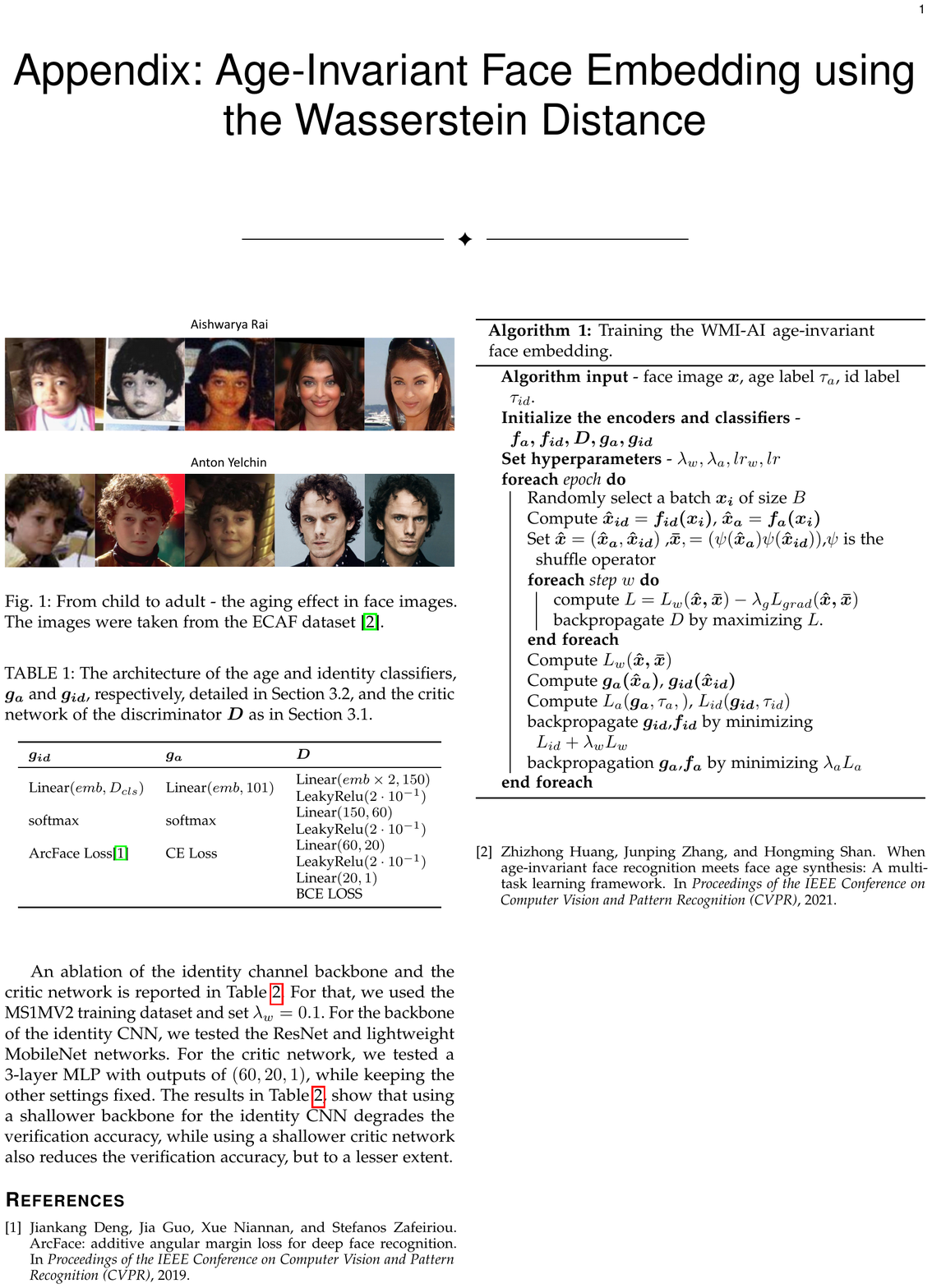}

\end{document}